\documentclass{article}

\usepackage[nonatbib, final]{neurips2024}

\usepackage[utf8]{inputenc} % allow utf-8 input
\usepackage[T1]{fontenc}    % use 8-bit T1 fonts
\usepackage{hyperref}       % hyperlinks
\usepackage{url}            % simple URL typesetting
\usepackage{booktabs}       % professional-quality tables
\usepackage{amsfonts}       % blackboard math symbols
\usepackage{nicefrac}       % compact symbols for 1/2, etc.
\usepackage{microtype}      % microtypography
\usepackage{xcolor}         % colors
\usepackage{blindtext}
\usepackage[numbers]{natbib}
\usepackage{ circuitikz }
\usepackage{tikz}
\usepackage{graphicx}
\usepackage{tabu}
\usepackage[]{caption}
\usepackage{authblk}
\title{
Climate Adaptation with Reinforcement Learning: Economic vs. Quality of Life Adaptation Pathways}

\author[1]{\textbf{Miguel Costa}$^{*,}$}
\author[1]{\textbf{Arthur Vandervoort}$^{*,}$}
\author[1]{\textbf{Martin Drews}}
\author[2]{\textbf{Karyn Morrissey}}
\author[1]{\textbf{Francisco C. Pereira}}

\affil[1]{Department of Technology, Management and Economics, 
Technical University of Denmark, 2800 Kgs. Lyngby, Denmark}
\affil[2]{J.E. Cairnes School of Business \& Economics, University of Galway, Galway, Ireland}

\affil[ ]{\texttt{\{migcos, apiva, mard, camara\}@dtu.dk}}
\affil[ ]{\texttt{karyn.morrissey@universityofgalway.ie}}

\begin{document}

\maketitle
\def\thefootnote{*}\footnotetext{These authors contributed equally to this work.}

\vspace{-18pt}
% % % % % % % % % % % % % % % % % % % % % % % % % % % % % % 
% % % % % %     Abstract
% % % % % % % % % % % % % % % % % % % % % % % % % % % % % % 
\begin{abstract}
\vspace{-8pt}
% Background, topic
Climate change will cause an increase in the frequency and severity of flood events, prompting the need for cohesive adaptation policymaking. Designing effective adaptation policies, however, depends on managing the uncertainty of long-term climate impacts. Meanwhile, such policies can feature important normative choices that are not always made explicit.
% Rationale for study, other research
We propose that Reinforcement Learning (RL) can be a useful tool to both identify adaptation pathways under uncertain conditions while it also allows for the explicit modelling (and consequent comparison) of different adaptation priorities (e.g. economic vs. wellbeing).
% Data, research, methods
We use an Integrated Assessment Model (IAM) to link together a rainfall and flood model, and compute the impacts of flooding in terms of quality of life (QoL), transportation, and infrastructure damage.
% Findings, significance
Our results show that models prioritising QoL over economic impacts results in more adaptation spending as well as a more even distribution of spending over the study area, highlighting the extent to which such normative assumptions can alter adaptation policy.
% Code
Our framework is publicly available: \url{https://github.com/MLSM-at-DTU/maat_qol_framework}.
\vspace{-8pt}
\end{abstract}

% % % % % % % % % % % % % % % % % % % % % % % % % % % % % % 
% % % % % %     Introduction
% % % % % % % % % % % % % % % % % % % % % % % % % % % % % % 
\section{Introduction}
\label{sec:introduction}
% ######################################################
% INTRODUCTIONS
% Flooding
As a changing climate continues to increase the likelihood of high-impact weather events, cities will have to adapt in order to safeguard the health and wellbeing of their residents \cite{ipcc2023climate}. Copenhagen, our case study, is one of Denmark's most  vulnerable cities in terms of social flood vulnerability \citep{prall_comprehensive_2024}, and is expected to experience an increase in the intensity and frequency of flood events \citep{olesen2014fremtidige}.

% what to optimize for
Pluvial urban flooding can severely disrupt transportation systems and the infrastructural, social, and economic networks that depend on them \cite{reballyFloodImpactAssessments2021a}. Consequently, policymakers must devise effective policies to mitigate these impacts. However, developing such policies requires contending with the complexity and uncertainty inherent in climate change and urban systems, as well as navigating difficult trade-offs between competing policy priorities. Policymakers may face decisions such as whether to prioritize economic or civil infrastructure protection, whether to implement interventions sooner or later, or whether to concentrate adaptations in few high-impact areas or many low-impact areas, all while addressing uncertainties regarding future flooding frequency and intensity.

% RL is great.
Climate adaptation policy development is therefore not a simple optimization problem. Rather, it requires models that explicitly account for the societal trade-offs inherent to adaptation strategies \citep{garner_climate_2016}. We suggest that artificial intelligence tools can address this challenge by serving as decision-support frameworks that integrate scientific evidence into policymaking \citep{tyler2023ai}. Reinforcement Learning (RL) in particular has the potential to help navigate these complex dynamics by identifying optimal policies that balance short- and long-term objectives, and the implications of different policy priorities \citep{gilbert2022choicesrisksrewardreports}.

% Objectives
In this work, we address the following research question: ``\textit{What climate adaptation pathways maximise quality of life (QoL), transport, and infrastructure-focused objectives, and what trade-offs exist between them?}'' To answer this question, we first develop an RL-based framework integrating rainfall projections, flood models, and transport simulation. Second, we use this framework to compare policies trained on different policy priorities, explicitly examining how policies prioritizing QoL, mobility, or infrastructure protection yield divergent adaptation pathways. Overall, this work demonstrates how AI-driven approaches can inform adaptation planning and provide actionable insights for policymakers addressing complex and competing demands of climate change adaptation.

% % % % % % % % % % % % % % % % % % % % % % % % % % % % % % 
% % % % % %     Framework
% % % % % % % % % % % % % % % % % % % % % % % % % % % % % % 
\section{Modelling Framework}
\label{sec:framework}
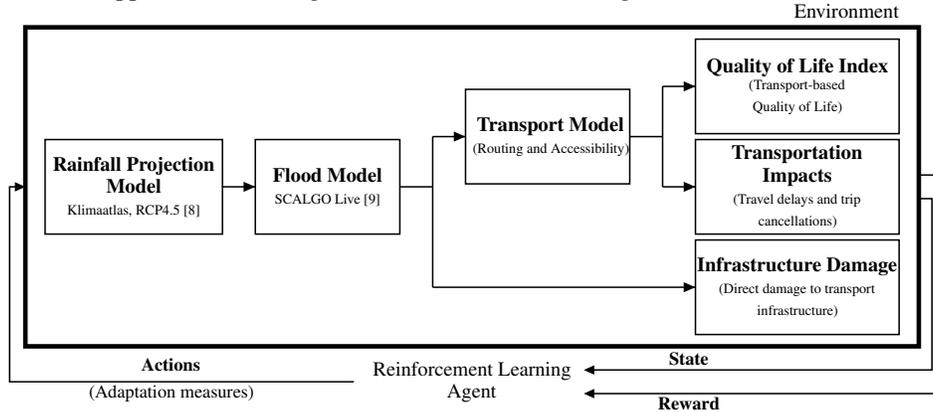
\begin{figure}[htb]
    \vspace{-12pt}
    \centering
    \resizebox{0.9\linewidth}{!}{%
        \tikzset{every picture/.style={line width=0.75pt}} %set default line width to 0.75pt        
        \begin{tikzpicture}[x=0.65pt,y=0.65pt,yscale=-.9,xscale=1]
                

%uncomment if require: \path (0,232); %set diagram left start at 0, and has height of 232

%Shape: Rectangle
\draw [line width=2.25] (25,25) -- (705,25) -- (705,295) -- (25,295) -- cycle  ;

% Text Node
\draw (365,325) node   [align=left] {\begin{minipage}[lt]{120pt}\setlength\topsep{0pt}
\begin{center}
Reinforcement Learning Agent
\end{center}
\end{minipage}};

%Straight Lines [id:da5897083012926433] 
\draw (705,170) -- (715,170) -- (715,315) -- (455,315) ;
\draw [shift={(450,315)}, rotate = 360] [fill={rgb, 255:red, 0; green, 0; blue, 0 }  ][line width=0.08]  [draw opacity=0] (8,-4) -- (0,0) -- (8,4) -- cycle ;
%Straight Lines [id:da6778049630294789] 
\draw (705,150) -- (724,150) -- (724,335) -- (455,335) ;
\draw [shift={(450,335)}, rotate = 360] [fill={rgb, 255:red, 0; green, 0; blue, 0 }  ][line width=0.08]  [draw opacity=0] (8,-4) -- (0,0) -- (8,4) -- cycle ;
%Straight Lines [id:da507302552582728] 
\draw (275,325) -- (13,325) -- (13,160) -- (20,160) ;
\draw [shift={(25,160)}, rotate = 180] [fill={rgb, 255:red, 0; green, 0; blue, 0 }  ][line width=0.08]  [draw opacity=0] (8,-4) -- (0,0) -- (8,4) -- cycle ;

% Text Node
\draw (530,305) node [font=\small] [align=left] {\textbf{State}};
% Text Node
\draw (530,343) node [font=\small] [align=left] {\textbf{Reward}};
% Text Node
\draw (136,310) node [font=\small] [align=left] {\begin{minipage}[lt]{32.83pt}\setlength\topsep{0pt}
\begin{center}
\textbf{Actions}
\end{center}
\end{minipage}};
% Text Node
\draw (136,335) node [font=\small] [align=left] {\begin{minipage}[lt]{100.pt}\setlength\topsep{0pt}
\begin{center}
{\footnotesize (Adaptation measures)}
\end{center}

\end{minipage}};
% Text Node
\draw (650,11) node [align=left] {Environment};

% Text Node
\draw    (40,120) -- (175,120) -- (175,200) -- (40,200) -- cycle  ;
\draw (107.5,160) node   [align=left] {\begin{minipage}[lt]{80pt}\setlength\topsep{0pt}
\begin{center}
\textbf{Rainfall Projection Model}\\{\scriptsize{Klimaatlas, RCP4.5 \citep{dmi2023klimaatlas}}}
\end{center}
\end{minipage}};

% Text Node
\draw    (200,120) -- (310,120) -- (310,200) -- (200,200) -- cycle  ;
\draw (255,160) node   [align=left] {\begin{minipage}[lt]{80pt}\setlength\topsep{0pt}
\begin{center}
\textbf{Flood Model}\\
{\scriptsize{SCALGO Live \citep{scalgo}}}
\end{center}
\end{minipage}};

% Connection
%\draw    (310,160) -- (335,160) ;

% Text Node
\draw    (360,77.5) -- (485,77.5) -- (485,157.5) -- (360,157.5) -- cycle  ;
\draw (425,117.5) node   [align=left] {\begin{minipage}[lt]{80pt}\setlength\topsep{0pt}
\begin{center}
\textbf{Transport Model}\\
{\scriptsize{(Routing and Accessibility)}}
\end{center}
\end{minipage}};

% Text Node
\draw    (535,35) -- (690,35) -- (690,115) -- (535,115) -- cycle  ;
\draw (612.5,75) node   [align=left] {\begin{minipage}[lt]{99.08pt}\setlength\topsep{0pt}
\begin{center}
\textbf{Quality of Life Index}\\
{\scriptsize{(Transport-based\\Quality of Life)}}
\end{center}
\end{minipage}};

% Text Node
\draw    (535,120) -- (690,120) -- (690,200) -- (535,200) -- cycle  ;
\draw (612.5,160) node   [align=left] {\begin{minipage}[lt]{99.08pt}\setlength\topsep{0pt}
\begin{center}
\textbf{Transportation Impacts}\\
{\scriptsize{(Travel delays and trip cancellations)}}
\end{center}
\end{minipage}};

% Text Node
\draw    (535,205) -- (690,205) -- (690,285) -- (535,285) -- cycle  ;
\draw (612.5,245) node   [align=left] {\begin{minipage}[lt]{99.08pt}\setlength\topsep{0pt}
\begin{center}
\textbf{Infrastructure Damage}\\
{\scriptsize{(Direct damage to transport infrastructure)}}
\end{center}
\end{minipage}};

% Connection
\draw    (175,160) -- (200,160) ;
\draw [shift={(200,160)}, rotate = 180] [fill={rgb, 255:red, 0; green, 0; blue, 0 }  ][line width=0.08]  [draw opacity=0] (8.93,-4.29) -- (0,0) -- (8.93,4.29) -- cycle ;
% Connection
\draw    (335,117.5) -- (360,117.5) ;
\draw [shift={(360,117.5)}, rotate = 180] [fill={rgb, 255:red, 0; green, 0; blue, 0 }  ][line width=0.08]  [draw opacity=0] (8.93,-4.29) -- (0,0) -- (8.93,4.29) -- cycle ;
% Connection
\draw    (510,75) -- (535,75) ;
\draw [shift={(535,75)}, rotate = 180] [fill={rgb, 255:red, 0; green, 0; blue, 0 }  ][line width=0.08]  [draw opacity=0] (8.93,-4.29) -- (0,0) -- (8.93,4.29) -- cycle ;
% Connection
\draw    (510,160) -- (535,160) ;
\draw [shift={(535,160)}, rotate = 180] [fill={rgb, 255:red, 0; green, 0; blue, 0 }  ][line width=0.08]  [draw opacity=0] (8.93,-4.29) -- (0,0) -- (8.93,4.29) -- cycle ;
% Connection
\draw    (335,245) -- (535,245) ;
\draw [shift={(535,245)}, rotate = 180] [fill={rgb, 255:red, 0; green, 0; blue, 0 }  ][line width=0.08]  [draw opacity=0] (8.93,-4.29) -- (0,0) -- (8.93,4.29) -- cycle ;

% Connection
\draw    (510,75) -- (510,160) ;
% Connection
\draw    (335,117.5) -- (335,245) ;

% Connection
\draw    (485,117.5) -- (510,117.5) ;
% Connection
\draw    (310,160) -- (335,160) ;

%\end{figure}
        \end{tikzpicture}
    }
    \vspace{-6pt}
    \caption{Integrated Assessment Model using RL to learn the best sequence of transport-related adaptation policies for rainfall events in Copenhagen from 2023-2100.}
    \label{fig:iam_framework}
\end{figure}

We frame our approach as an Integrated Assessment Model (\autoref{fig:iam_framework}), which we outline below.

\vspace{-12pt}

% ######################################################
% RAINFALL PROJECTION MODEL
% ######################################################
\subsection{Rainfall Projection Model and Flood Model}

We begin by sampling rainfall event intensities (amount of precipitation) from probability distributions built from  \citep{dmi2023klimaatlas} for 2023--2100. Based on the distribution we sampled one rainfall event per year. For simplicity and as proof-of-concept, we assumed the rainfall intensity to be equal to the accumulated daily rainfall. For each rainfall event, we modelled the associated urban flood using SCALGO Live \citep{scalgo}. Using this approach, water is distributed according to precipitation intensity and terrain properties. Both nuisance and heavy precipitation-induced flooding events are simulated, which allowed for a range of water depths to be included.

% ######################################################
% TRANSPORTATION MODEL
% ######################################################
\subsection{Transportation Model}

For the transportation component, we perform two transport simulations: routing and accessibility.

\vspace{-6pt}\paragraph{Trip Routing:}
We began by downloading the drivable, cyclable, and walkable street network data from OpenStreetMap (OSM) \citep{osm2024} using \texttt{osmnx} \citep{boeing2024modeling}. Next, we divided Copenhagen's Inner city in 29 Traffic Analysis Zones (TAZs) following the Danish National Transport Model \citep{vejdirektoratet2022gmm}. In each TAZ, origin and destination (OD) points were randomly assigned to simulate trips within and across TAZs. Finally, we simulated car, bicycle, and on foot trips following trips' distribution in Copenhagen's inner city \citep{christiansen2024tu}. For each trip, we simulated its shortest travel time path route using STL \citep{koehler2025stable}. Naturally, water level in each road, cycling lane, or sidewalk impacts travel time. As water levels rise, travel speed decreases, until, at a certain level, travel becomes impossible. In turn, travel time increases as individuals travel at a reduced speed or because they need to avoid a flooded segment. For accounting for water depth effects on travel, we modelled its impacts via depth-disruption functions \citep{pregnolato2017impact, finnis2008field}.

\vspace{-6pt}\paragraph{Accessibility:}
We started by subdividing Copenhagen into a hex grid using \texttt{h3} \citep{h32025uber}. Then, we extracted different categories of points of interest (POIs) from OSM following \cite{dobrowolska_mapping_2024}. Using the center point of each hex, we computed how accessible available POIs were based on computing the shortest path from each hex grid centroid to that location. In this work, we employed a cumulative accessibility approach, i.e., we counted the number of locations that are reachable within a set travel time for each transport mode (10-minute walk, a 15-minute cycle, or a 30-minute drive).

% ######################################################
% ECONOMIC IMPACTS
% ######################################################
\subsection{Transport-related Impacts}

We model flood impacts on transportation via three pathways: direct infrastructure damage, indirect increases in travel time and trip cancellations, and in reduced quality of life.

\vspace{-6pt}\paragraph{Direct Infrastructure Impacts:}
Transport infrastructure is directly vulnerable to urban flooding. We began by estimating a transport network link's (i.e., road, cycling lane, sidewalk) total construction costs depending on its road type, number of lanes, presence of light posts, and traffic lights \citet{ginkel2021flood}. Then, using  depth-damage functions \citep{ginkel2021flood}, we estimated the percentage of damage depending on the water depth at that location, as it's associated economic damage. This damage accounts for reconstruction, repair, cleaning, and resurfacing works needed to restore the link to its original state. Finally, we aggregate these monetary losses at the $i$-th TAZ as $I_{i}$.

\vspace{-6pt}\paragraph{Indirect Transport Impacts:}
Next, we estimated floods' indirect impacts on travel via two types of impacts: delays caused by increased travel times and cancellations when no traversable path between OD existed. First, we computed travel delays by comparing simulated trips' travel time for a given rainfall event and a no-rain scenario. Travel delays were then modelled as economic losses using the Danish travel delays value of time (VoT) \citep{transportministeriet2022enhedspriser}, which were then aggregated at the TAZ-level as $D_{i}$. Second, when no path was available for a trip due to high water levels, we considered that trip as cancelled and modelled its economic loss as 80\% of the VoT as typically considered for cancelled trips \citep{hallenbeck2014travel}. Finally, we aggregated cancelled trips economic losses at the $i$-th TAZ as $C_{i}$.

\vspace{-6pt}\paragraph{Quality of Life Impacts:}
QoL impacts are based on an index measuring the per-capita number of accessible services and amenities for each location in our study area \cite{dobrowolska_mapping_2024}. We computed the index by taking the number of POIs accessible from a hex (together with the downweighted sum for the hex's neighbours' counts) and divided by the hex's population. Counts were then normalised at the 75th percentile for each POI category, while counts that exceed the 75th percentile were clipped at 1. Per-category QoL estimates were combined using a weighted sum, with weights determined using a logistic regression model predicting urban life satisfaction based on satisfaction with amenities from \citep{ec/dgregioQualityLifeEuropean2023}. This yields an index ranging between 1 (accessibility greater than or equal to the 75th percentile for each category) and 0 (no POIs accessible). Finally, QoL was aggregated at the TAZ-level as $Q_i$.

% ######################################################
% REINFORCEMENT LEARNING
% ######################################################
\subsection{Reinforcement Learning}
We posit to learn the best sequence of adaptation measures using RL. RL uses an agent-based approach that interacts with an environment by taking actions to maximise a (delayed) reward function \citep{sutton2018reinforcement}.

In this work, we propose eight actions (adaptation measures) that can be implemented in each area of Copenhagen. These include different real-world solutions for increasing road drainage or water storage. Actions directly change the environment and both directly and indirectly affect our estimated impacts. We define the reward function to optimize for as:\vspace{-3pt}
\begin{equation}
    R = \sum_{i} 
    \beta_{I} I_{i} +
    \beta_{D} D_{i} +
    \beta_{C} C_{i} +
    \beta_{Q} Q_{i} + 
    \beta_{A} A_{i} +
    \beta_{M} M_{i}\vspace{-3pt}
\end{equation}
where $I_{i}$, $D_{i}$, $C_{i}$, and $Q_{i}$ are as defined earlier, $A_{i}$ is the cost of applying an action, and $M_{i}$ its maintenance cost. This function is highly customisable, allowing users to choose $\beta$ weights depending on competing priorities between economic losses, quality of life, and economic costs of actions.

To learn which action to perform, our RL agent takes an action and collects information on the state of our digital city. Over time, the agent learns to balance trade-offs between competing actions and input uncertainty and the best sequence of policies that maximise the cumulative reward.

% % % % % % % % % % % % % % % % % % % % % % % % % % % % % % 
% % % % % %     RESULTS 
% % % % % % % % % % % % % % % % % % % % % % % % % % % % % % 
\section{Preliminary Experiments \& Discussion}
\label{sec:results}
% ######################################################
% EXPERIMENTAL SETUP
We setup our IAM using Python, Gymnasium interface \citep{towers_gymnasium_2023}, Stable-Baseline3 \citep{stable_baselines3}, and PPO \citep{huang2020closer, schulman2017proximal}. As preliminary experiments, we present results for Copenhagen's inner city (29 TAZ) and set the time horizon between 2023--2100. Here, we showcase results for two model typologies: EC (Economic Costs), where we focus adapting entirely based on economic damages ($\beta_I=\beta_D=\beta_C=\beta_A=\beta_M=1; \beta_Q=0$); and QoL, where we focus on optimising based on QoL and disregard other economic impacts ($\beta_Q=0.5; \beta_A=\beta_M=1; \beta_I=\beta_D=\beta_C=0$).\footnote{For this experiment we report results based on $\beta_Q = 0.5$. Future research will explore the implications of setting $\beta_Q$ to different values.}

% The experiments were run for five distinct seed runs to allow for different weather projections and robustness. Unless stated otherwise, we present results for these runs.

% ######################################################
% RESULTS
\begin{figure}[!tb]
  \centering
  %%%%% action distribution
  % \begin{minipage}{.50\textwidth}
  %   \centering
  %   \includegraphics[trim={0cm 6cm 0cm 0cm},clip, width=.98\textwidth]{figs/action_distribution_ec.png} 
  % \end{minipage}%
  % \begin{minipage}{.50\textwidth}
  %   \centering
  %   \includegraphics[trim={0cm 6cm 0cm 0cm},clip, width=.98\textwidth]{figs/action_distribution_qol.png} 
  % \end{minipage}%

  %%%%% spending plots
  \begin{minipage}[b]{\textwidth}
    \centering
    \includegraphics[width=.9\linewidth]{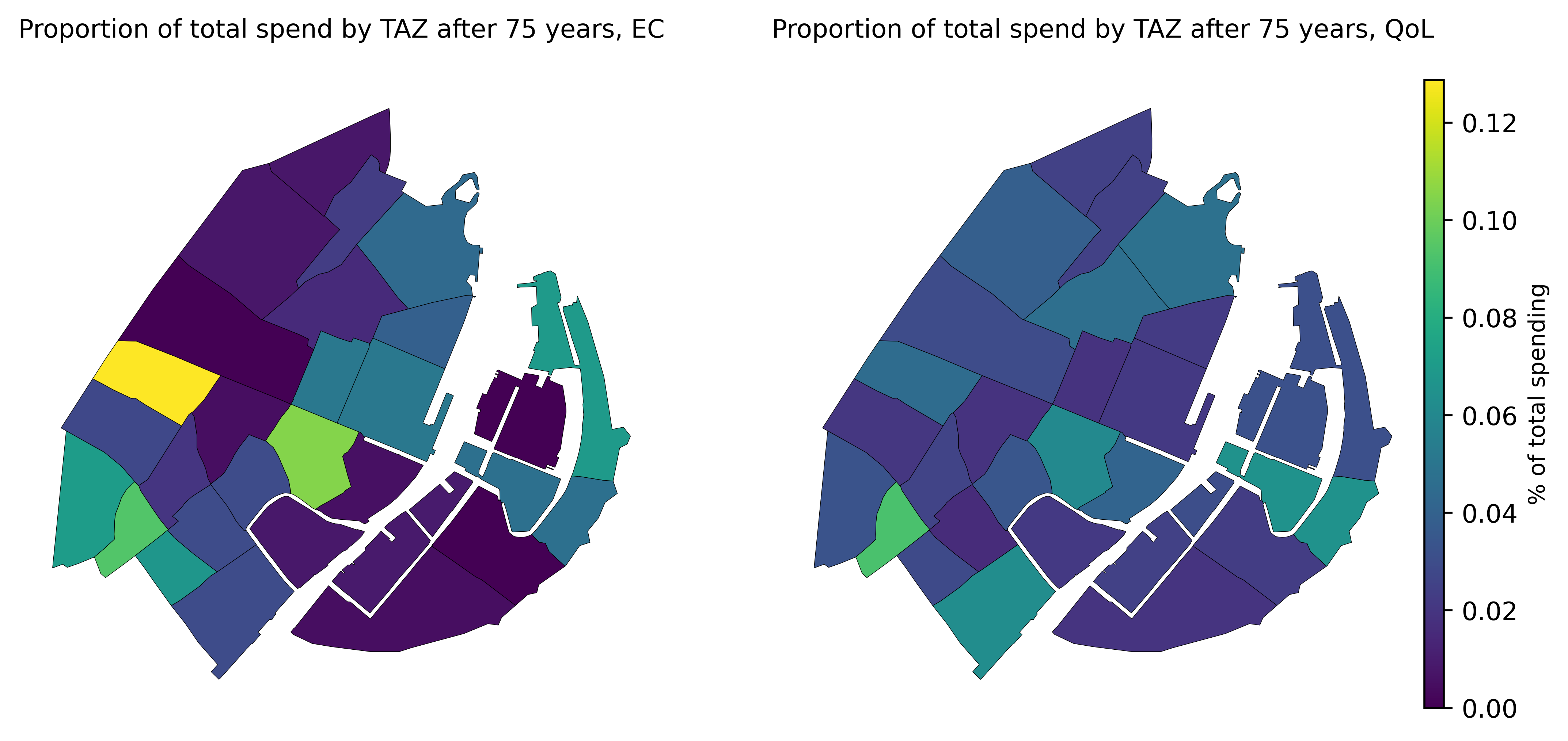}
  \end{minipage}

  \vspace{2pt}
    
  \begin{minipage}{.50\textwidth}
    \centering
    \includegraphics[trim={0cm 0cm 5.5cm .9cm},clip, width=.98\textwidth]{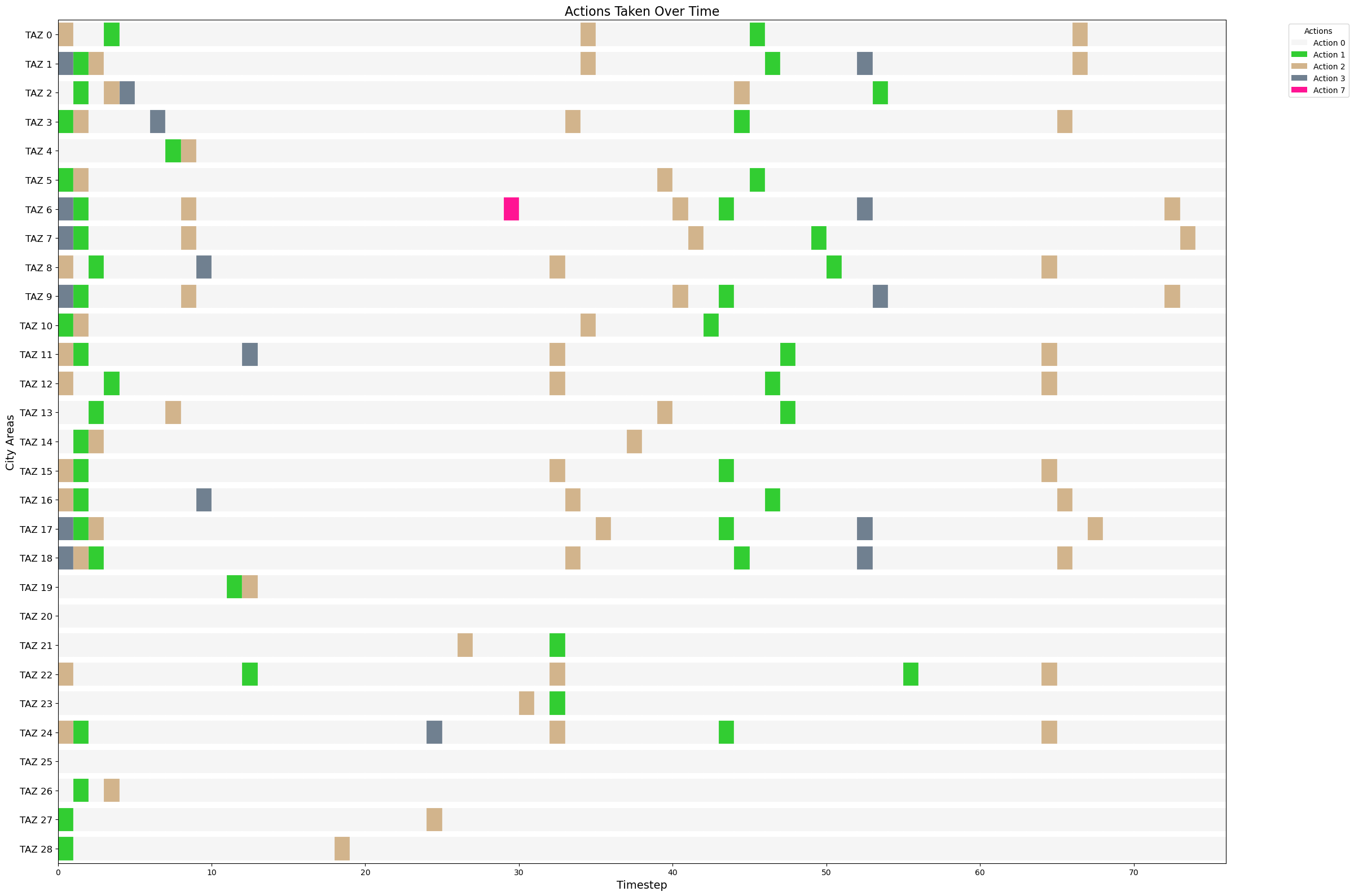} 
  \end{minipage}%
  \begin{minipage}{.50\textwidth}
    \centering
    \includegraphics[trim={0cm 0cm 5.5cm .8cm},clip, width=.98\textwidth]{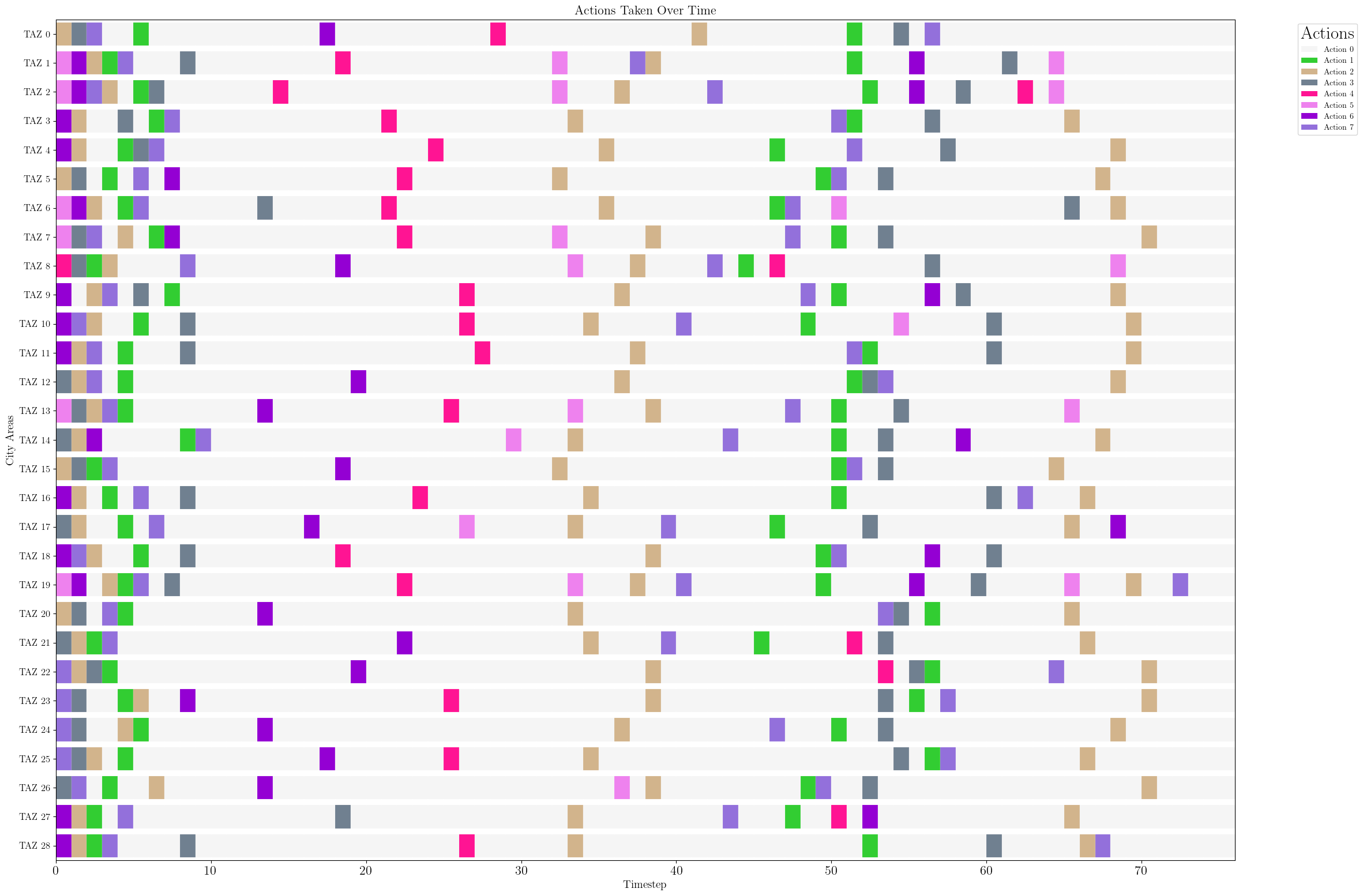} 
  \end{minipage}%

  \begin{minipage}{.50\textwidth}
    \centering
    \includegraphics[trim={0cm 0cm 0cm 15cm},clip, width=.98\textwidth]{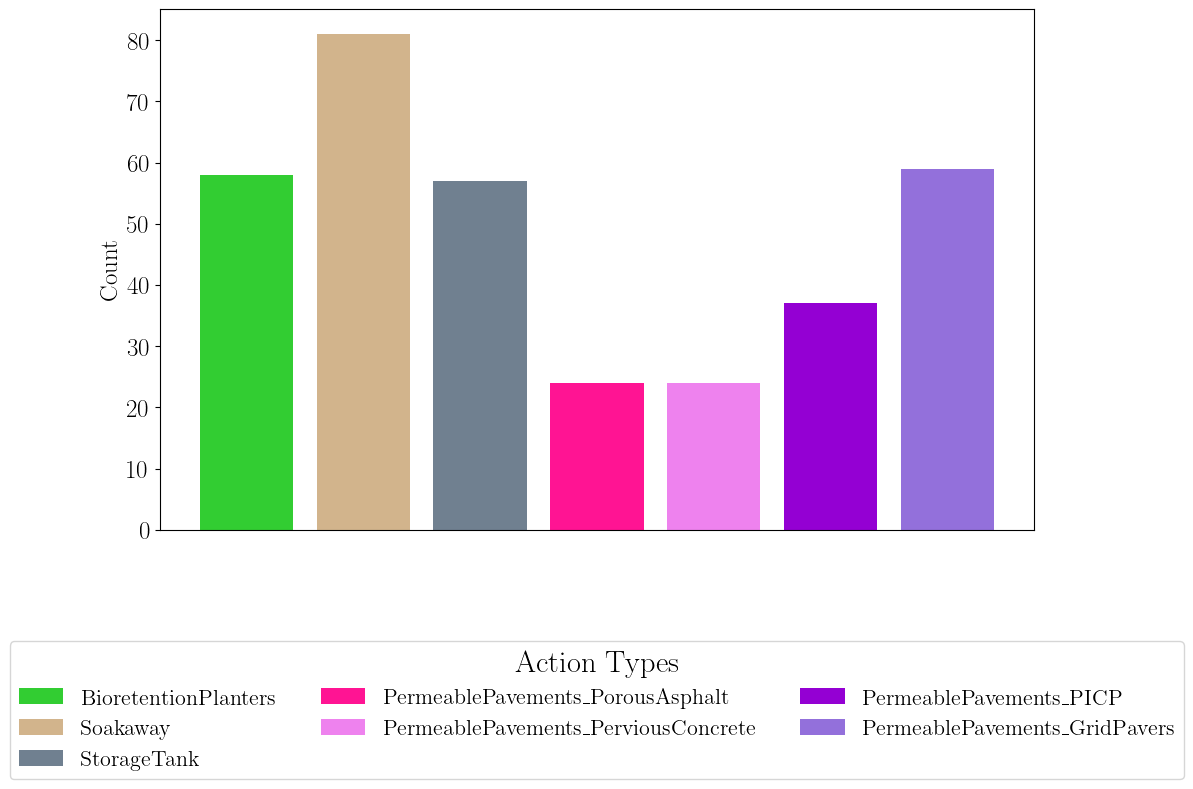} 
  \end{minipage}%
  
\caption{\textbf{Top row:} The proportion of money spent by TAZ by 2100 for the EC and QoL policies. \textbf{Bottom row:} Adaptation pathways per TAZ for the EC (left) and the QoL policies (right).}
\label{fig:actions}
\vspace{-10pt}
\end{figure}

% #######################
% Present results for EC. Then present results for QOL.
Our main results are shown in Figure ~\ref{fig:actions}. 
First, for the EC-focused model, the RL agent' main goal is to balance between when and where to invest in adaptation measures to minimise the expected flood-related impacts in infrastructure damage, travel delays and trip cancellations. Results show that the agent takes about 0.103 adaptation measures per year until 2100 with a total spending of DKK 2.1bn. ($\sim$€280mn.). In contrast, the RL agent in the QoL-focused model applies many more actions in its effort to minimise QoL losses and adaptation costs. Here, we notice that the average number of adaptation measures is 0.559 per year, totalling in DKK 20bn. ($\sim$€2.7bn.) spent over our 75-year study period.

% #######################
% Compare the results for both policies
There are three important takeaways from these results. First, climate adaptation planning requires different strategies when optimizing purely for economic returns or when including quality of life dimensions. Second, the QoL-focused policy distributed its spending more evenly over the study area and used a wider range of actions when compared to the EC-focused policy. Third and finally, regardless of the focus of each model, the proposed framework showcases that it can be a valuable tool balancing short- and long-term objectives and navigate deep uncertainty in climate change adaptation planning.

% ######################################################
% DISCUSSION
Overall, our results contribute to calls in the adaptation literature to develop the means to explicitly model the societal trade-offs inherent to adaptation policy \citep{garner_climate_2016}, while also highlighting that wellbeing-focused adaptation policies can produce meaningfully different policy pathways compared to more financially focused policies \citep{quinn_health_2023}. Taken together, our results demonstrate the feasibility of using RL to support long-term adaptation planning as well as the trade-offs different policies might entail. In the future we expect to expand our framework to the full city of Copenhagen, explore distributional (socio-demographic) effects of policies, and explore model specifications based on insights from the transportation justice literature (e.g., Foster-Greer-Thorbecke indices \citep{karner_advances_2024}).

% % % % % % % % % % % % % % % % % % % % % % % % % % % % % % 
% % % % % %     Acknowledgements
% % % % % % % % % % % % % % % % % % % % % % % % % % % % % % 
\newpage
\begin{ack}
This work was supported by a research grant (VIL57387) from VILLUM FONDEN.
\end{ack}

% % % % % % % % % % % % % % % % % % % % % % % % % % % % % % 
% % % % % %     References
% % % % % % % % % % % % % % % % % % % % % % % % % % % % % % 
\bibliographystyle{IEEEtranN}
\bibliography{references}

% % % % % % % % % % % % % % % % % % % % % % % % % % % % % % 
% % % % % %     Appendices
% % % % % % % % % % % % % % % % % % % % % % % % % % % % % % 
%\newpage
%\appendix

%\section{Appendix / supplemental material}

%Optionally include supplemental material (complete proofs, additional experiments and plots) in appendix.
%All such materials \textbf{SHOULD be included in the main submission.}

\end{document}